%% file: main.tex
\documentclass{article} 

\input{configs/math_commands}

\usepackage{iclr2027_conference}
\usepackage{times}
\usepackage[hypertexnames=false]{hyperref}
\usepackage{url}
\usepackage{algorithm}
\usepackage{algpseudocode}
\usepackage{booktabs}
\usepackage{multirow}
\usepackage{graphicx}
\usepackage{xspace}
\usepackage{enumitem}
\usepackage{bbm}
\usepackage{pifont}
\usepackage{comment}
\usepackage[table]{xcolor}

\newcommand{\sysname}{\textsc{TraceQuery}\xspace}

\newif\ifshowcomments
\showcommentstrue   
\ifshowcomments
    \newcommand{\sg}[1]{\textcolor{orange}{\textbf{[Shangqian: #1]}}}
\else
    \newcommand{\sg}[1]{}
\fi

\ifshowcomments
    \newcommand{\xl}[1]{\textcolor{orange}{\textbf{[Xin: #1]}}}
\else
    \newcommand{\xl}[1]{}
\fi

\newcommand{\warn}[1]{\textcolor{red}{}}

\title{\sysname: Learning What to Recompute from Full-Denoising Trajectory in Diffusion Language Models}

\input{metadata/author}

\iclrfinalcopy 
\begin{document}

\maketitle

\input{sections/0_abstract}
\input{sections/1_introduction}
\input{sections/2_relatedworks}
\input{sections/3_preliminaries}
\input{sections/4_methodology}
\input{sections/5_experiments}
\input{sections/6_conclusion}

\bibliography{bibs/references}
\bibliographystyle{bibs/iclr2027_conference}

\appendix
\input{sections/7_appendix}

\end{document}

%% file: configs/math_commands.tex
\usepackage{amsmath,amsfonts,bm}

\def\eqref#1{equation~\ref{#1}}

\def\1{\bm{1}}

\DeclareMathAlphabet{\mathsfit}{\encodingdefault}{\sfdefault}{m}{sl}
\SetMathAlphabet{\mathsfit}{bold}{\encodingdefault}{\sfdefault}{bx}{n}



%% file: metadata/author.tex
\author{
\makebox[\textwidth][c]{
\textbf{Yukai Jin}\textsuperscript{1},
\textbf{Zhankai Ye}\textsuperscript{1},
\textbf{Jiayue Yang}\textsuperscript{2},
\textbf{Fangyi Li}\textsuperscript{3}
}
\\
\makebox[\textwidth][c]{
\textbf{Shuoqiu Li}\textsuperscript{1},
\textbf{Xu Gao}\textsuperscript{4},
\textbf{Shibo Li}\textsuperscript{1},
\textbf{Xin Liu}\textsuperscript{1},
\textbf{Shangqian Gao}\textsuperscript{1}
}
\\[0.5ex]
\makebox[\textwidth][c]{
\textsuperscript{1}Florida State University
\quad
\textsuperscript{2}Carnegie Mellon University
\quad
\textsuperscript{3}University of Pennsylvania
\quad
\textsuperscript{4}ByteDance Inc.
}
}

%% file: sections/0_abstract.tex
\begin{abstract}

Diffusion large language models (dLLMs) enable flexible generation through iterative masked denoising, but incur substantial inference cost because token representations are repeatedly recomputed across denoising steps.
Existing cache-based methods reduce this cost by reusing intermediate results across denoising steps and recomputing only a subset of positions at each step, chosen either on a fixed schedule or by heuristic signals of change. Our trajectory experiments instead point to decoding trajectory as the more relevant quantity. Cached generation remains accurate when it stays close to the full-denoising trajectory, with errors concentrated around unmask-order deviations and reduced as longer reference prefixes are recovered. This suggests a more direct principle: \textbf{limiting representation error helps, but keeping generation on the full-denoising trajectory is a more direct route to accurate decoding}.
We therefore formulate efficient dLLM inference as \emph{trajectory-guided querying}.
Rather than broadly repairing representation error, we use full-denoising trajectories to supervise which unresolved positions should receive limited fresh computation. Based on this formulation, we introduce \sysname, a lightweight sparse decoding framework that predicts a small set of near-future trajectory-relevant positions, queries only those candidates, and lets the backbone select which one to commit from their fresh predictions.
\sysname delivers the strongest quality–computation trade-off among cache-based approaches, sustaining high accuracy with substantially fewer fresh token computations, and generalizes across tasks, model variants, and families.
Code: \url{https://github.com/fish1928/tracequery.git}

\end{abstract}

%% file: sections/1_introduction.tex
\section{Introduction}
\label{Introduction}

Diffusion large language models (dLLMs) have emerged as a promising alternative to autoregressive language models by replacing fixed left-to-right generation with iterative masked denoising~\citep{diff9_mdlm,diff10_radd,diff11_llada}.
By repeatedly predicting unresolved tokens under bidirectional attention, dLLMs support flexible generation orders and expose opportunities for parallel decoding~\citep{diff12_dream,diff13_seed,diff14_mercury}.
However, these advantages come with substantial inference cost.
At each denoising step, standard dLLMs repeatedly recompute token representations over large portions of the sequence, even though only one or a few tokens may ultimately be committed ~\citep{kv_diff2_flashdlm,kv_diff5_entropycache}.
When the number of denoising steps grows with generation length, this redundant recomputation can lead to cubic decoding cost~\citep{kv_diff1_dkv,kv_diff4_dllmcache}. To reduce this redundancy, recent work maintains an approximate KV cache across denoising steps, reusing stable representations and selectively refreshing only a subset of positions~\citep{kv_diff1_dkv,kv_diff3_fastdllm,kv_diff4_dllmcache,kv_diff6_d2cache}. Despite increasingly fine-grained refresh policies, these methods share a common objective: \textit{limiting the representation error introduced by cache reuse}.

We revisit this objective by examining how cached decoding behaves relative
to full denoising. Our trajectory analysis (Figure~\ref{fig:diagram2}) shows
that cached decoding deviates from the full-denoising unmasking order even
when token-level predictions agree. Accuracy also drops as the injected
full-denoising prefix shortens, tying quality to keeping generation anchored
to the reference trajectory. These observations motivate the following causal
view:
\begin{center}
\emph{representation error} $\rightarrow$ \emph{decision perturbation} $\rightarrow$ \emph{trajectory divergence} $\rightarrow$ \emph{quality degradation}.
\end{center}
Refreshing cache entries intervenes at the first link of this chain by reducing representation error, but it controls decision perturbation and trajectory divergence only indirectly. The link from representation error to decoding decisions is highly non-uniform. Many errors are decision-irrelevant and leave the next commitment unchanged, while a small subset can alter a consequential decision and redirect the subsequent trajectory. 
This makes representation repair inefficient, because computation may be spent correcting harmless errors, while even useful corrections must propagate through later representations and decisions before improving the decoding trajectory.  This suggests a more direct principle: \textbf{limiting representation error helps, but keeping generation on the full-denoising trajectory is a more direct route to accurate decoding.}

\input{figures/2_trajectory_test/2_trajectory_test_tex}

This view gives the full-denoising trajectory a new role.
A full-denoising run provides a sequence of decisions made by the original backbone from fully recomputed representations, without cache-induced error.
Although multiple decoding trajectories may lead to equally good generations, this trajectory is immediately available from the model itself and requires neither additional trajectory search nor external supervision. It therefore provides a natural, unapproximated reference for anticipating the decoding decisions that lie in the near future.

Based on this view, we formulate efficient dLLM inference as \emph{trajectory-guided querying}.
Rather than searching for an optimal trajectory, we use the already available full-denoising trajectory to guide where a limited query budget should be spent.
At each denoising step, the objective is to identify unresolved positions that appear soon along the full-denoising reference trajectory and include them in a small fresh-query set.
Importantly, the trajectory does not determine the final decoding decision.
It supervises candidate selection, while the backbone queries the selected candidates with fresh computation and retains control over which position to commit.

We introduce \textbf{\sysname}, a lightweight trajectory-guided sparse decoding framework illustrated in Figure~\ref{fig:diagram3}.
During training, \sysname learns from full-denoising trajectories to rank unresolved positions according to their near-future relevance.
During inference, it selects a small candidate set from inexpensive signals available in the current decoding state.
Only these candidates receive fresh computation, while the remaining positions continue to reuse cached representations.
The backbone dLLM then selects the final commitment from the freshly evaluated candidates.
\sysname therefore splits the two decisions: the router decides where fresh computation goes, and the backbone decides what to commit.

Our main contributions are summarized as follows:

\begin{itemize}[leftmargin=*, labelsep=0.4em, itemsep=4pt, topsep=4pt, parsep=0pt]
    
\item We propose \emph{trajectory-guided querying}. Representation error degrades generation through perturbed consequential decisions, so rather than spending limited computation to repair the error, we let the available full-denoising trajectory guide it directly toward the decisions that lie ahead.
    
\item We build \textbf{\sysname}, a lightweight sparse decoding framework that implements this formulation with a router trained on full-denoising trajectories. From inexpensive decoding-state signals, the router decides which few positions receive fresh computation, and the backbone decides what to commit from the freshly evaluated candidates.
    
\item A single learned router approaches full-denoising quality with substantially fewer fresh token computations across tasks, model variants, and dLLM families, improving the quality--computation trade-off over cache-based baselines.

\end{itemize}

%% file: figures/2_trajectory_test/2_trajectory_test_tex.tex
\begin{figure}[t]
    \centering
    \includegraphics[width=\linewidth]{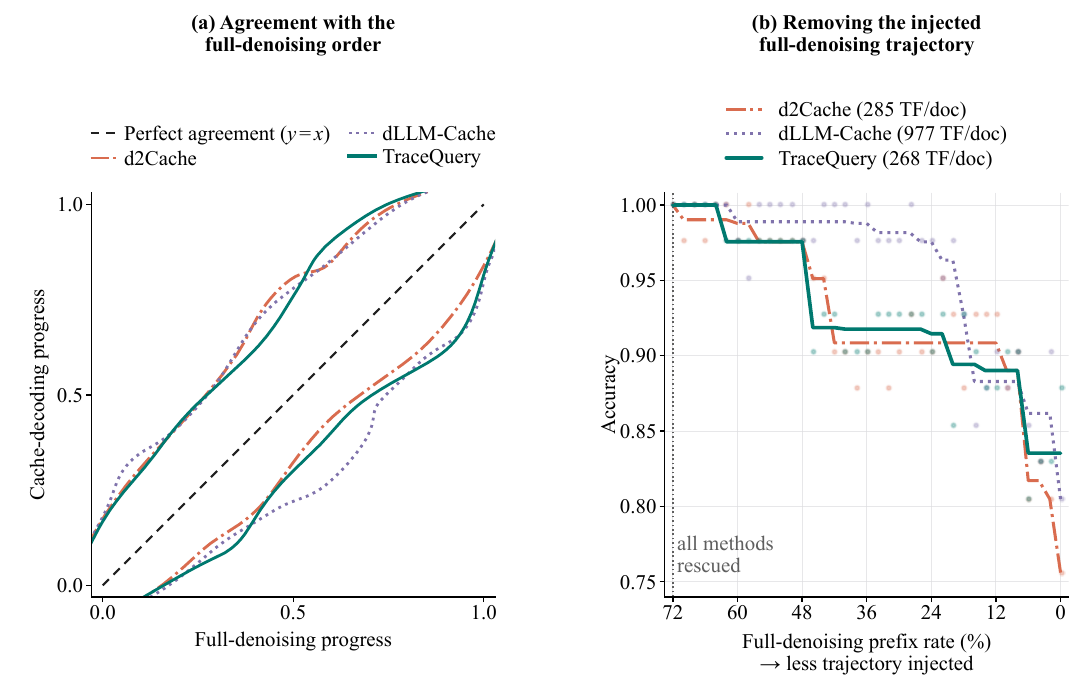}
    \caption{
    \textbf{Trajectory agreement and dependence on trajectory guidance.}
    \textbf{(a)} We compare decoding progress under full denoising and cached decoding. Each underlying observation corresponds to a token position, paired by the step at which it is unmasked under full denoising and its corresponding step under cached decoding. Rather than plotting individual observations, we show the 80\% density contours. Concentration along the diagonal indicates agreement with the full-denoising unmask order, while broader deviations reflect trajectory divergence.
    \textbf{(b)} Accuracy on examples solved by full denoising as the injected
    full-denoising prefix is progressively removed. 
    }
    \label{fig:trajectory-agreement}
    \label{fig:diagram2}
\end{figure}

%% file: sections/2_relatedworks.tex
\vspace{-5 pt}
\section{Related Work}
\vspace{-5 pt}
\label{sec:related_work}

Diffusion language models generate text through iterative masked denoising rather
than fixed left-to-right factorization, enabling bidirectional context and
flexible decoding orders~\citep{diff3_d3pm,diff4_diffusionlm,diff8_sedd,
diff9_mdlm,diff10_radd}. Recent large-scale models such as LLaDA and Dream
demonstrate that this paradigm can scale to competitive language modeling and
reasoning performance~\citep{diff11_llada,diff12_dream}, while systems such as
Seed and Mercury further highlight its potential for parallel generation
~\citep{diff13_seed,diff14_mercury}.

\textbf{Learned and Trajectory-Aware Decoding.}
Prior methods learn, search, guide, or verify unmasking decisions and
denoising trajectories~\citep{decoding_unmask_policy,decoding_lookahead_path,
kv_diff2_flashdlm,decoding_trajectory_spec}. \sysname instead learns which
masked positions to \emph{query}: its router allocates fresh computation,
while the backbone determines the committed position and token.
Full-denoising trajectories provide offline supervision, with no external
guider, online search, or draft verification at inference.

\textbf{KV Reuse and Adaptive Refresh.}
Exact KV reuse is difficult in dLLMs because bidirectional attention changes
token representations as decoding proceeds. Early methods reuse states across
completed blocks~\citep{kv_diff0_block} or stabilized regions, with periodic
recomputation to limit staleness~\citep{kv_diff2_flashdlm,kv_diff3_fastdllm}.
dKV~\citep{kv_diff1_dkv} identifies \emph{local fidelity}: each decoding
decision changes representations most strongly at a limited set of positions,
making uniform refresh unnecessary. Building on this principle,
dLLM-Cache~\citep{kv_diff4_dllmcache} uses V-state verification to select
refresh positions, while d2Cache~\citep{kv_diff6_d2cache} combines
confidence and local mask density for unresolved positions with
attention-based selection for prompt and decoded positions. These methods
adapt which cached states to refresh; \sysname focuses on which unresolved
positions receive fresh queries before the next commitment.

Different from the above works, our analysis suggests that the downstream effect of cache-induced error is more closely reflected by trajectory deviation than by representation error alone. We use the available full-denoising trajectory to guide fresh computation toward positions that are likely to matter for upcoming decoding decisions.

%% file: sections/3_preliminaries.tex
\section{Preliminaries}
\label{sec:preliminaries}

This section introduces masked diffusion decoding, the full-denoising reference
trajectory, and the cache setting used by \sysname. These definitions apply to
both single-block and block-wise diffusion decoding.

\subsection{Masked Diffusion Decoding}
\label{sec:trajectory}

A masked diffusion language model operates on a partially masked sequence
$x^{(t)}$ at denoising step $t$. Let $\mathcal{M}_t$ and $\mathcal{U}_t$
denote the unresolved and already-resolved positions, respectively. The model
predicts token distributions for positions in $\mathcal{M}_t$ in parallel, and
a decoding policy selects a subset
$\mathcal{A}_t \subseteq \mathcal{M}_t$ to commit:
\begin{equation}
    p_{\mathrm{DLLM}}(x_{1:N}\mid\tau)
    =
    \prod_{t=0}^{T-1}
    \prod_{i\in\mathcal{A}_t}
    p_{\theta}(x_i \mid x^{(t)}),
    \qquad
    \tau =
    (\mathcal{A}_0,\ldots,\mathcal{A}_{T-1}),
\end{equation}
where $N$ is the sequence length, $T$ is the number of denoising steps, and $\tau$ denotes the decoding
trajectory induced by the sequence of commitments. After committing $\mathcal{A}_t$, the resolved and unresolved sets evolve as
\begin{equation}
    \mathcal{U}_{t+1}
    =
    \mathcal{U}_t \cup \mathcal{A}_t,
    \qquad
    \mathcal{M}_{t+1}
    =
    \mathcal{M}_t \setminus \mathcal{A}_t.
\end{equation}
Under single-token decoding, $\mathcal{A}_t=\{i_t\}$, where $i_t$ is the
position selected at step $t$. Unlike autoregressive decoding, $t$ specifies
only temporal progression and is not tied to a fixed sequence position.
Consequently, the ordering of committed positions itself becomes part of the
decoding process.

\subsection{Full-Denoising Reference Trajectory}
\label{sec:reference_trajectory}

For a given prompt, we fix the backbone model, decoding policy, denoising
schedule, and token-sampling configuration. At each denoising step, the
full-denoising baseline performs a forward pass over the entire active
generation region and recomputes the representations used to make the next
commitment. In particular, it does not reuse approximate representations from
earlier denoising steps. The decoding policy then selects the next position
from predictions obtained using these fully recomputed representations.

Let $\mathcal{A}_t^{\mathrm{FD}}$ denote the set of positions committed at
step $t$ by full denoising. Under single-token decoding,
$\mathcal{A}_t^{\mathrm{FD}} = \{i_t^{\mathrm{FD}}\}$, and we define the
full-denoising reference trajectory as
\begin{equation}
    \tau^{\mathrm{FD}}
    =
    \left(
        \mathcal{A}_0^{\mathrm{FD}},
        \ldots,
        \mathcal{A}_{T-1}^{\mathrm{FD}}
    \right)
    =
    \left(
        i_0^{\mathrm{FD}},
        \ldots,
        i_{T-1}^{\mathrm{FD}}
    \right).
    \label{eq:full_denoising_trajectory}
\end{equation}
The trajectory records only the order in which positions are unmasked; it does
not include the identities of the committed tokens. The superscript
$\mathrm{FD}$ indicates that the corresponding decisions are produced by full
denoising. We use $\tau^{\mathrm{FD}}$ as an operational reference for the behavior of the
unmodified backbone and decoding policy in the absence of cache-induced
approximation. It is neither assumed to be the unique decoding trajectory nor
guaranteed to produce a correct or optimal output. Nevertheless, it can be
generated offline directly from the frozen backbone, without external
annotations or additional trajectory search, and therefore provides a
well-defined target for supervising and evaluating trajectory agreement under
cached decoding.

\subsection{Cached Diffusion Inference}
\label{sec:generation_region}

We partition the sequence into three regions: the prefix $\mathcal{P}$, the
active generation region $\mathcal{G}$, and the postfix $\mathcal{S}$.
Previously completed blocks belong to $\mathcal{P}$, while subsequent blocks
belong to $\mathcal{S}$. At denoising step $t$, the active generation region
is partitioned into
\begin{equation}
    \mathcal{G}
    =
    \mathcal{U}_t \cup \mathcal{M}_t,
    \qquad
    \mathcal{U}_t \cap \mathcal{M}_t = \emptyset.
\end{equation}
Decoding starts with a full forward pass to initialize the persistent KV cache.
Following existing region-wise caching strategies
~\citep{kv_diff3_fastdllm,kv_diff4_dllmcache}, the active generation region is
periodically recomputed to control accumulated approximation error, while the
prefix and postfix are reused over longer intervals. \sysname operates only
within the active generation region $\mathcal{G}$.

%% file: sections/4_methodology.tex
\section{Methodology}
\label{sec:methodology}

\input{figures/3_main_process/3_main_process_tex}

\subsection{Trajectory Recovery under a Query Budget}
\label{sec:trajectory_recovery}

Let $\tau^{\mathrm{FD}}$ be the full-denoising reference trajectory defined in
Section~\ref{sec:reference_trajectory}. After step $t$, its remaining ordered
suffix is
\begin{equation}
    \pi_t^{\mathrm{FD}}
    =
    \left(
        i_{t+1}^{\mathrm{FD}},
        \ldots,
        i_{T-1}^{\mathrm{FD}}
    \right).
    \label{eq:future_reference_trajectory}
\end{equation}
Given a maximum query horizon $H$, we define
\begin{equation}
\begin{aligned}
    H_t
    &=
    \min\!\left(
        H,
        \left|\mathcal{M}_{t+1}^{\mathrm{FD}}\right|
    \right), \\
    \pi_{t,H_t}^{\mathrm{FD}}
    &=
    \left(
        i_{t+1}^{\mathrm{FD}},
        \ldots,
        i_{t+H_t}^{\mathrm{FD}}
    \right), \\
    \mathcal{Q}_{t+1}^{\mathrm{oracle}}
    &=
    \left\{
        i_{t+1}^{\mathrm{FD}},
        \ldots,
        i_{t+H_t}^{\mathrm{FD}}
    \right\},
\end{aligned}
\label{eq:future_reference_prefix}
\end{equation}
where $\mathcal{M}_{t+1}^{\mathrm{FD}}$ is the masked set in the corresponding
full-denoising state. The ordered prefix
$\pi_{t,H_t}^{\mathrm{FD}}$ specifies the next $H_t$ reference decisions,
while $\mathcal{Q}_{t+1}^{\mathrm{oracle}}$ is the query set that would be
available if the reference trajectory were known at inference time. Here,
``oracle'' denotes access to $\tau^{\mathrm{FD}}$, rather than global
optimality. We formulate computation allocation as predicting this near-future reference
set from the current decoding state. During training,
$\pi_{t,H_t}^{\mathrm{FD}}$ provides ordered supervision; during inference,
\sysname selects at most $H$ positions using only signals available from the
cached decoding state. Thus, the target is determined by future decoding
behavior rather than by cached-state magnitude, age, or representation error.

\subsection{Trajectory-Supervised Router Training}
\label{sec:router_training}

We train the \sysname router offline on intermediate states collected from
full-denoising trajectories. For each reference state, let
$i_{t,j}^{\mathrm{FD}}$ denote the $j$-th element of
$\pi_{t,H_t}^{\mathrm{FD}}$.

For every unresolved position
$i\in\mathcal{M}_{t+1}^{\mathrm{FD}}$, we construct
\begin{equation}
\begin{aligned}
    \mathbf{z}_i^{(t)}
    &=
    \left[
        a_{i,\mathrm{last}}^{(t)},
        \rho_i^{(t)},
        d_i^{(t)}
    \right], \\
    s_i^{(t)}
    &=
    f_\phi\!\left(\mathbf{z}_i^{(t)}\right),
\end{aligned}
\label{eq:router_features_and_score}
\end{equation}
where the features are normalized last-layer attention from recently committed
positions, local mask density, and relative distance to recently committed
positions, respectively. The scorer $f_\phi$ is a two-layer MLP shared across
positions and denoising steps. Detailed feature definitions are provided in
Appendix~\ref{app:signal_definitions}.

At ranking stage $j$, previously selected reference positions are removed:
\begin{equation}
    \mathcal{C}_{t,j}
    =
    \mathcal{M}_{t+1}^{\mathrm{FD}}
    \setminus
    \left\{
        i_{t,1}^{\mathrm{FD}},
        \ldots,
        i_{t,j-1}^{\mathrm{FD}}
    \right\}.
    \label{eq:ranking_candidate_set}
\end{equation}
We train the router with a truncated Plackett--Luce objective:
\begin{equation}
    \mathcal{L}_{\mathrm{PL}}
    =
    -
    \frac{1}{|\mathcal{D}|}
    \sum_{(x,t)\in\mathcal{D}}
    \sum_{j=1}^{H_t}
    \log
    \frac{
        \exp\!\left(s_{i_{t,j}^{\mathrm{FD}}}^{(t)}\right)
    }{
        \displaystyle
        \sum_{i\in\mathcal{C}_{t,j}}
        \exp\!\left(s_i^{(t)}\right)
    },
    \label{eq:router_training_objective}
\end{equation}
where $\mathcal{D}$ is the collection of intermediate reference states. The
objective assigns higher scores to positions that occur earlier in the future
full-denoising order. The reference trajectory is used only during offline
training.

\subsection{\sysname Inference}
\label{sec:tracequery_inference}

At inference time, the trained router scores every unresolved position
$i\in\mathcal{M}_{t+1}$ and selects the $H_t$ highest-scoring positions:
\begin{equation}
    \mathcal{Q}_{t+1}
    =
    \underset{
        \substack{
            \mathcal{Q}\subseteq\mathcal{M}_{t+1} \\
            |\mathcal{Q}|=H_t
        }
    }{\operatorname{arg\,max}}
    \sum_{i\in\mathcal{Q}} s_i^{(t)}.
    \label{eq:query_set_selection}
\end{equation}
The router scores determine only which positions are queried. For each
$i\in\mathcal{Q}_{t+1}$, the backbone recomputes its representation against
the persistent KV cache and obtains a query-updated token distribution
$\widetilde{p}_i^{(t+1)}$. Because the remaining representations are cached,
this sparse query is not equivalent to full recomputation.

Under confidence-based greedy single-token decoding, the backbone computes
\begin{equation}
\begin{aligned}
    \widehat{v}_i^{(t+1)}
    &=
    \underset{v\in\mathcal{V}}{\operatorname{arg\,max}}\;
    \widetilde{p}_i^{(t+1)}(v), \\
    c_i^{(t+1)}
    &=
    \widetilde{p}_i^{(t+1)}
    \left(\widehat{v}_i^{(t+1)}\right), \\
    i_{t+1}
    &=
    \underset{i\in\mathcal{Q}_{t+1}}{\operatorname{arg\,max}}\;
    c_i^{(t+1)}, \qquad
    \mathcal{A}_{t+1}=\{i_{t+1}\}.
\end{aligned}
\label{eq:greedy_query_commitment}
\end{equation}
It replaces the mask at $i_{t+1}$ with
$\widehat{v}_{i_{t+1}}^{(t+1)}$ and updates
\begin{equation}
    \mathcal{M}_{t+2}
    =
    \mathcal{M}_{t+1}
    \setminus
    \{i_{t+1}\}.
    \label{eq:masked_set_update}
\end{equation}
All other positions in
$\mathcal{Q}_{t+1}\setminus\{i_{t+1}\}$ remain masked, while positions outside
$\mathcal{Q}_{t+1}$ continue to reuse their cached representations. The loop
repeats between scheduled broad-recomputation steps until the active generation
region is resolved. No reference trajectory or oracle computation is used at
inference time.

%% file: figures/3_main_process/3_main_process_tex.tex
\begin{figure}[t]
    \centering
    \includegraphics[width=\textwidth]{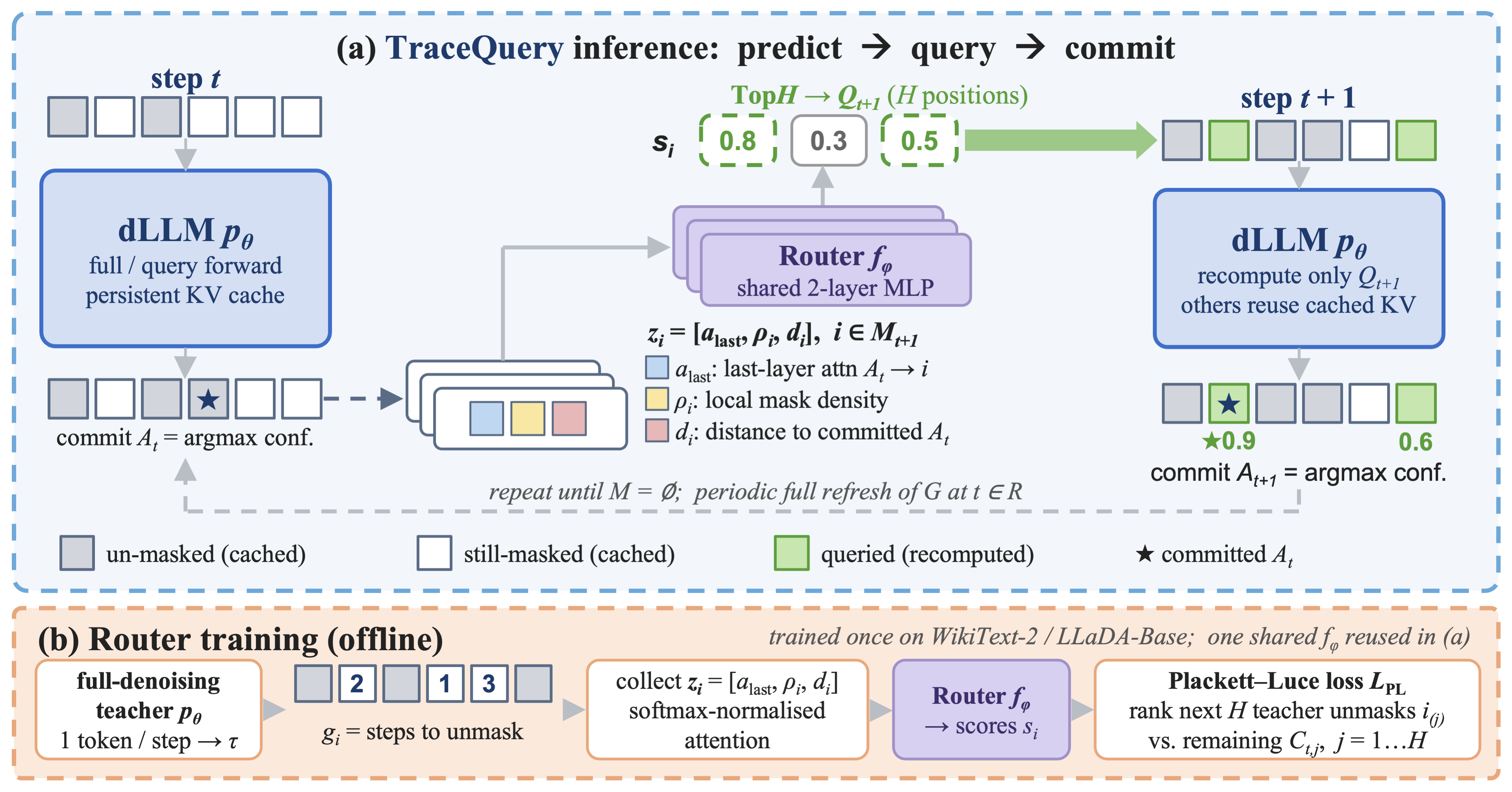}
    \caption{
    Overview of \sysname.
    Full denoising first provides an offline reference trajectory describing the
    future order of decoding decisions. The router is trained to recover the
    near-future portion of this trajectory from inexpensive signals available
    during decoding. At inference time, after step $t$ commits
    $\mathcal{A}_t$, the router ranks the remaining unresolved positions and
    selects a small query set $\mathcal{Q}_{t+1}$. Only these positions receive
    fresh computation, while other positions reuse cached representations. The
    backbone dLLM then makes the final commitment from the freshly queried
    candidates. Periodic broad recomputation is retained to control accumulated
    approximation error.
    }
    \label{fig:diagram3}
\end{figure}

%% file: sections/5_experiments.tex
\section{Experiments}
\label{sec:experiments}

We have shown in Figure~\ref{fig:diagram2} that trajectory fidelity
is strongly associated with downstream performance. Based on this
observation, we ask three questions: whether trajectory-guided querying yields
end-to-end gains across models and tasks, how efficiently it allocates
computation relative to existing cache-based methods, and which router design
choices account for its performance.

\subsection{Experimental Setup}
\label{sec:exp_setup}

\textbf{Models.} We evaluate four representative diffusion language models:
LLaDA-Base, LLaDA-Instruct, Dream-Base, and Dream-Instruct.
For the instruction-tuned models, we do not apply chat templates.
In our preliminary evaluation, chat templates improve performance on some
benchmarks but degrade it on others, introducing an additional source of
variation unrelated to the inference method itself. We therefore use a unified
prompt format across all models.

\textbf{Benchmarks.} We evaluate reasoning, knowledge, and code generation using five benchmarks:
MATH-500(512), BBH-125(256), GSM8K(256), HumanEval(512), and MBPP(512). All benchmarks use the same precision, decoding configuration, and hardware environment.

\textbf{Baselines.} We compare \sysname against full denoising and two selective cache-based
acceleration methods:  dLLM-Cache, and d2Cache.
The controlled comparison varies the operating point of each method to obtain
matched-quality and matched-computation settings, while the final end-to-end
evaluation uses the adopted configuration of each method.

\textbf{Efficiency metrics.} We report downstream task score together with total computation in TFLOPs,
an wall-clock inference time. For batch experiments, we
additionally report the number of samples that can be processed under the same resource budget. TFLOPs measure the algorithmic computation footprint, while wall-clock time captures the practical execution cost of each method.

\subsection{End-to-End Evaluation}
\label{sec:e2e}
\input{tables/5_end_to_end}
We begin with the main result: end-to-end performance across four backbone
models and five benchmarks. Each method uses its adopted operating
configuration. We compare full denoising,
dLLM-Cache, d2Cache, and \sysname for each backbone.

Table~\ref{tab:e2e} reports downstream task performance together with total
computation, and wall-clock inference time. Full denoising
provides the quality reference, while the three cache-based methods represent
different strategies for reducing repeated computation through approximate
reuse and selective refresh.

Across the four backbones, TraceQuery trades a modest drop in accuracy for substantially lower compute: it retains 90-98\% of full-denoising accuracy while using roughly 14× fewer TFLOPs per document, the lowest among the caching methods compared. Compared with the selective
cache-based baselines, the advantage is particularly pronounced in TFLOPs and
wall-clock time, consistent with the small query footprint observed in the
controlled comparison. The improvement is also maintained across both base and
instruction-tuned models, suggesting that the learned querying strategy is not
specific to a single model variant or task family.

The controlled comparison in Section~\ref{sec:controlled_efficiency}
examines how computation allocation contributes to these results.

\subsection{Controlled Efficiency Comparison}
\label{sec:controlled_efficiency}

To examine this allocation directly, we compare \sysname against
dLLM-Cache and d2Cache under controlled resource budgets. This experiment varies the operating point of each method to
isolate how efficiently fresh computation is allocated. We consider two
complementary settings: matched quality and matched computation.

\input{figures/6_horizontal_comparison/6_horizontal_comparison_v4}

\textbf{Matched-quality computation.} Figure~\ref{fig:diagram6}(a) compares the amount of computation required to
reach the same downstream quality. The x-axis reports total computation
in TFLOPs. For each
method, we adjust its computation budget and evaluate the resulting average
performance across the all benchmarks.

At the same quality level, \sysname requires substantially less fresh
computation than other selective cache-based baselines. This difference follows from the
allocation objective. dLLM-Cache, and d2Cache spend computation
maintaining or refreshing a comparatively large set of cached representations,
whereas \sysname uses a small query horizon and concentrates fresh computation
on candidates expected to matter for upcoming decoding decisions. As a result,
\sysname reaches the same quality with a smaller computation footprint.

\paragraph{Matched-compute batch efficiency.}
We next study how the per-sample query footprint affects batched inference.
For sample $b$, let $\mathcal{Q}_t^{(b)}$ denote the positions requiring fresh
computation at denoising step $t$. Batched execution combines these positions
into a shared query set:
$
\mathcal{U}_t(B)
=
\bigcup_{b=1}^{B}\mathcal{Q}_t^{(b)}.
$
As batch size increases, different samples contribute different queried
positions, causing this union to grow. Methods with a large per-sample refresh
budget can therefore quickly cover most of the generation region, making sparse
batched inference increasingly close to dense recomputation.

In contrast,
\sysname uses a much smaller query horizon, so the union grows substantially
more slowly and sparse computation remains effective at larger batch sizes. Under the same computation budget, \sysname therefore supports substantially
larger batches than the competing cache methods. The advantage is modest in the
single-sample setting but grows rapidly with batch size, showing that reducing
the per-sample query footprint provides an additional systems benefit beyond
reducing FLOPs alone.

Together, the two controlled comparisons show that trajectory-guided querying
improves both \emph{quality efficiency}, by achieving the same quality with less
computation, and \emph{batch efficiency}, by accommodating more samples under
the same resource budget.

\subsection{Ablation Study}
\label{sec:ablation}

We then examine the router design underlying these results through the
two-stage ablation shown in Figure~\ref{fig:diagram4}.
All results are averaged across LLaDA-Base,
LLaDA-Instruct, Dream-Base, and Dream-Instruct.

\paragraph{Stage I: Router design.}

\input{figures/4_ablation_test/4_ablation_test_tex}

We tune the router on full-denoising trajectories, varying one design choice at a time and measuring Recall@5 over 17,670 held-out positions. Features matter most: adding geometry and confidence raises Recall@5 from 0.51 (attention alone) to 0.61 and 0.78, respectively. Stale confidence loses 13 points, while nonlinear architectures differ by less than 0.6 points. Loss and normalization choices are similarly stable, and performance holds for \(H=3\)–\(12\); we adopt \(H=5\). We train on mixed tasks and retrain for each backbone. Stage II tests whether confidence remains useful under cached inference.

\paragraph{Stage II: Closed-loop feature evaluation.}
Stage I evaluates the router under offline trajectory supervision, but some
features may behave differently once the model operates with cached states.
We therefore fix all design choices selected in Stage I and evaluate alternative
feature groups under closed-loop cached decoding.

For each benchmark, we sample 100 prompts and measure accuracy relative to the adopted router. We test whether confidence and margin remain useful when cached, since training observes fresh values but inference updates them only on re-query. We compare four feature sets: \textbf{Clean} uses only attention and geometry; \textbf{Policy-aged} carries values forward according to the deployment refresh schedule; \textbf{Random-aged} draws values from earlier steps; and \textbf{Conf/margin + age} adds the time since each position’s last refresh.
Closed-loop results reverse the offline ranking. We adopt Clean: conf/margin + age changes accuracy by $-3.8$ points (CI includes zero), while random-aged and policy-aged significantly reduce it by $5.2$ and $7.8$ points, respectively. Confidence is highly predictive offline, yet every cached variant underperforms without it, making closed-loop evaluation essential for feature selection.

%% file: tables/5_end_to_end.tex
\begin{table*}[!t]
\centering
\caption{
\textbf{End-to-end quality and inference efficiency.}
We evaluate four dLLM backbones on five downstream benchmarks with batch size 16. Each method uses its adopted operating configuration. TPS denotes tokens per second.
}
\label{tab:e2e}
\resizebox{\textwidth}{!}{
\begin{tabular}{llcccccc|ccc}
\toprule
Model
& Method
& MATH-500
& BBH-125
& GSM8K
& MBPP
& HumanEval
& Avg.$\uparrow$
& TFLOPs $\downarrow$
& TPS $\uparrow$
& Speedup $\uparrow$ \\
\midrule

\multirow{4}{*}{\shortstack{LLaDA\\Base}}
& Full Denoising & 30.1 & 44.7 & 70.7 & 39.6 & 32.9 & 100.0 & 5931 & 53 & $1.0\times$ \\
\cmidrule{2-11}
& dLLM-Cache & 29.5 & 43.5 & 70.2 & 37.8 & 32.9 & 98.1 & 1605 & 65 & $1.23\times$ \\
& d2Cache & 29.0 & 44.9 & 68.5 & 38.8 & 29.3 & 96.6 & 804 & 33 & $0.6\times$ \\
& \cellcolor{blue!10}\sysname
& \cellcolor{blue!10}28.1
& \cellcolor{blue!10}39.7
& \cellcolor{blue!10}67.0
& \cellcolor{blue!10}34.2
& \cellcolor{blue!10}29.3
& \cellcolor{blue!10}91.0
& \cellcolor{blue!10}415
& \cellcolor{blue!10}112
& \cellcolor{blue!10}$2.1\times$ \\
\midrule

\multirow{4}{*}{\shortstack{LLaDA\\Instruct}}
& Full Denoising & 34.9 & 58.0 & 76.3 & 36.8 & 35.4 & 100.0 & 5828 & 53 & $1.0\times$ \\
\cmidrule{2-11}
& dLLM-Cache & 34.3 & 56.4 & 73.4 & 28.4 & 36.6 & 94.9 & 1590 & 65 & $1.23\times$ \\
& d2Cache & 34.9 & 40.6 & 52.8 & 35.5 & 36.6 & 83.0 & 794 & 33 & $0.6\times$ \\
& \cellcolor{blue!10}\sysname
& \cellcolor{blue!10}33.7
& \cellcolor{blue!10}49.1
& \cellcolor{blue!10}75.1
& \cellcolor{blue!10}35.4
& \cellcolor{blue!10}35.4
& \cellcolor{blue!10}94.7
& \cellcolor{blue!10}409
& \cellcolor{blue!10}112
& \cellcolor{blue!10}$2.1\times$ \\
\midrule

\multirow{4}{*}{\shortstack{Dream\\Base}}
& Full Denoising & 39.8 & 43.2 & 71.6 & 55.0 & 37.8 & 100.0 & 5554 & 56 & $1.0\times$ \\
\cmidrule{2-11}
& dLLM-Cache & 39.1 & 39.2 & 74.1 & 54.6 & 34.8 & 97.7 & 1252 & 84 & $1.5\times$ \\
& d2Cache & 38.2 & 41.1 & 72.4 & 55.6 & 38.4 & 99.3 & 753 & 36 & $0.6\times$ \\
& \cellcolor{blue!10}\sysname
& \cellcolor{blue!10}37.2
& \cellcolor{blue!10}40.3
& \cellcolor{blue!10}72.0
& \cellcolor{blue!10}52.0
& \cellcolor{blue!10}30.5
& \cellcolor{blue!10}93.8
& \cellcolor{blue!10}388
& \cellcolor{blue!10}119
& \cellcolor{blue!10}$2.1\times$ \\
\midrule

\multirow{4}{*}{\shortstack{Dream\\Instruct}}
& Full Denoising & 41.3 & 49.2 & 79.2 & 52.8 & 49.4 & 100.0 & 5457 & 56 & $1.0\times$ \\
\cmidrule{2-11}
& dLLM-Cache & 41.8 & 58.3 & 83.7 & 54.6 & 45.7 & 104.5 & 1243 & 84 & $1.5\times$ \\
& d2Cache & 40.0 & 47.2 & 75.3 & 53.0 & 46.6 & 96.4 & 743 & 36 & $0.6\times$ \\
& \cellcolor{blue!10}\sysname
& \cellcolor{blue!10}38.6
& \cellcolor{blue!10}50.1
& \cellcolor{blue!10}80.6
& \cellcolor{blue!10}50.0
& \cellcolor{blue!10}47.6
& \cellcolor{blue!10}98.1
& \cellcolor{blue!10}387
& \cellcolor{blue!10}119
& \cellcolor{blue!10}$2.1\times$ \\
\bottomrule

\end{tabular}
}
\end{table*}

%% file: figures/6_horizontal_comparison/6_horizontal_comparison_v4.tex
\begin{figure}[h]
    \centering
    \begin{minipage}[h]{0.49\textwidth}
        \centering
        \includegraphics[width=\linewidth]{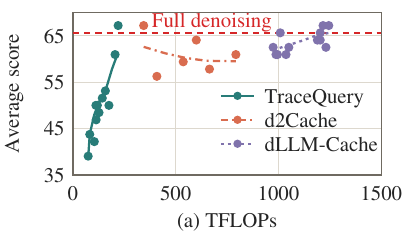}
    \end{minipage}\hfill
    \begin{minipage}[h]{0.49\textwidth}
        \centering
        \includegraphics[width=\linewidth]{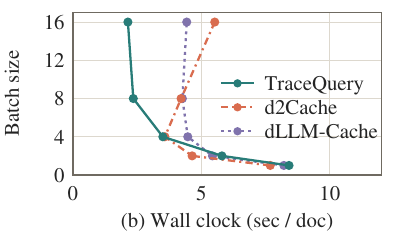}
    \end{minipage}
    \caption{
    \textbf{Quality--compute and batch-scaling comparison.}
    \textbf{(a)} Average score across benchmarks as a function of total
    computation, measured in TFLOPs. The red dashed line marks the score of
    full denoising without caching. \sysname reaches higher scores under
    comparable compute budgets.
    \textbf{(b)} Supported batch size as a function of wall-clock time per
    document. \sysname supports larger batches at comparable latency,
    demonstrating improved batch-scaling efficiency.
    }
    \label{fig:diagram6}
\end{figure}

%% file: figures/4_ablation_test/4_ablation_test_tex.tex
\begin{figure}[h]
    \centering
    \includegraphics[width=\textwidth]{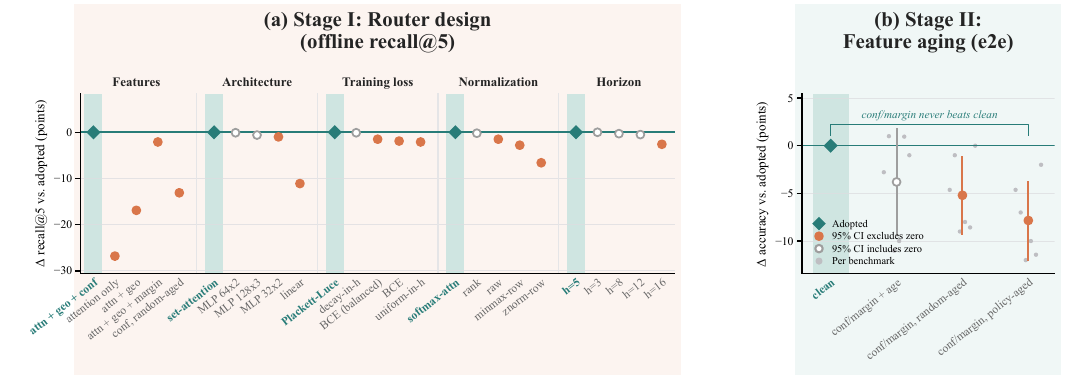}
    \caption{
    \textbf{Two-stage ablation of \sysname.}
    (a) Stage I selects the router configuration by varying the feature combination, router architecture, ranking loss, normalization, and query horizon during router training.
    (b) Stage II fixes the selected configuration and evaluates alternative feature groups under closed-loop cached decoding.
    Results are averaged across four models and reported as accuracy change relative to the adopted configuration. Bars show paired 95\% bootstrap confidence intervals; filled markers indicate intervals excluding zero, and
    gray dots denote per-benchmark changes.
    }
    \label{fig:diagram4}
\end{figure}

%% file: sections/6_conclusion.tex
\section{Conclusion}
\label{sec:conclusion}

We formulate efficient dLLM inference as \emph{trajectory-guided querying}:
allocating limited computation to positions likely to matter for upcoming
decoding decisions. \sysname learns this allocation from full-denoising
trajectories, then uses inexpensive decoding-state signals to select a small
query set at inference time. The backbone's updated predictions determine
which position and token to commit. Across LLaDA and Dream models, \sysname
improves the quality--computation trade-off over cache-based baselines, with
further gains in batched inference from its small per-sample query footprint.

\subsection*{AI use statement}

In this work, we used generative AI tools only for language polishing, grammar correction, improving the clarity of author-written text, and identifying potentially relevant related work. We did not use generative AI tools to generate research ideas, formulate mathematical claims, design experiments, implement methods, or analyze results. All AI-assisted content, including suggested related work, was reviewed, verified, and edited by the authors. We take responsibility for the final content of this work, including all text, claims, and artifacts.

%% file: sections/7_appendix.tex
\section{Detailed Computation and Memory Analysis}
\label{app:complexity}

\paragraph{Computation Complexity}

Let $S$ denote the total sequence length visible to attention, $B$ the size of
the active generation region, $T$ the number of denoising steps, $H$ the query
horizon, $R$ the number of full refresh steps, and $L$ the number of
Transformer layers.

Under single-token decoding, a regular sparse step recomputes at most
\begin{equation}
    q = H + 1
\end{equation}
positions, consisting of $H$ future-query candidates and one recently
committed position required by delayed cache admission. Near the end of a
generation block, $q$ may be smaller when fewer than $H$ masked positions
remain.

Querying $q$ positions against a KV cache of length $S$ requires
$\mathcal{O}(LqS)$ attention computation, whereas refreshing all $B$ positions
in the active generation region requires $\mathcal{O}(LBS)$. Ignoring the
one-time cache initialization, the total attention cost is therefore
\begin{equation}
    C_{\text{\sysname}}
    =
    \mathcal{O}
    \left(
        LS
        \left[
            RB + (T-R)(H+1)
        \right]
    \right).
    \label{eq:sparsequery_complexity}
\end{equation}
By comparison, full recomputation at every denoising step requires
\begin{equation}
    C_{\mathrm{Full}}
    =
    \mathcal{O}(LSTB).
    \label{eq:full_complexity}
\end{equation}

For single-block diffusion, $B=T=N$, where $N$ denotes the generation length.
Assuming $S=\mathcal{O}(N)$ and treating $H$ and $R$ as bounded constants,
\begin{equation}
    C_{\text{\sysname}}
    =
    \mathcal{O}(LN^2),
    \qquad
    C_{\mathrm{Full}}
    =
    \mathcal{O}(LN^3).
\end{equation}
Thus, selective querying reduces the dominant attention-computation scaling
from cubic to quadratic in generation length.

\paragraph{Concrete Decoding Configurations}

Our default experiments use a query horizon of $H=5$. For single-block
generation with $N=256$, we perform $R=16$ global refreshes. A regular sparse
step therefore recomputes at most six positions: five future-query candidates
and one recently committed position. Using the number of attention query
positions as a computation proxy, the relative attention workload is
approximately
\begin{equation}
    \frac{
        16\times256 + (256-16)\times6
    }{
        256\times256
    }
    \approx 8.45\%.
\end{equation}
This quantity measures the reduction in attention-query computation and should
not be interpreted as wall-clock speedup. Practical runtime additionally
depends on projection and MLP computation, cache operations, router execution,
kernel efficiency, and hardware utilization.

For a length-$256$ generation divided into $16$ blocks, each active block
contains $16$ positions. We perform no additional periodic full refresh within
each block and use the same query horizon $H=5$. The effect of query horizon
and refresh schedule is evaluated empirically in
Section~\ref{sec:experiments}.

\paragraph{Persistent Memory Consumption}

\sysname maintains a persistent KV cache for the complete sequence. Let
$d_{\mathrm{KV}}$ denote the dimensionality of the stored KV representation
per token, $b$ the number of bytes per element, and $B_{\mathrm{batch}}$ the
batch size. The persistent KV-cache memory is
\begin{equation}
    M_{\mathrm{KV}}
    =
    2 B_{\mathrm{batch}} L S d_{\mathrm{KV}} b,
\end{equation}
where the factor of two accounts for keys and values. Thus,
\begin{equation}
    M_{\mathrm{KV}}
    =
    \mathcal{O}(LSd_{\mathrm{KV}}),
\end{equation}
which scales linearly with sequence length.

For illustration, assuming a BF16 LLaDA-8B configuration with $L=32$,
$d_{\mathrm{KV}}=4096$, and batch size $1$, the model parameters occupy
approximately
\begin{equation}
    M_{\mathrm{model}}
    \approx
    8\times10^9\times2
    =
    16~\mathrm{GB}.
\end{equation}
For a cached sequence of length $S=256$, the corresponding KV cache requires
\begin{equation}
    M_{\mathrm{KV}}
    =
    2\times32\times256\times4096\times2
    \approx
    128~\mathrm{MiB},
\end{equation}
corresponding to less than $1\%$ additional persistent storage relative to the
BF16 model weights.

\begin{table}[t]
    \centering
    \caption{
    Persistent KV-cache memory for the illustrative BF16 LLaDA-8B
    configuration under batch size 1. Relative memory denotes model weights
    plus KV cache divided by model-weight memory alone.
    }
    \label{tab:kv_memory}
    \begin{tabular}{ccc}
        \toprule
        Cached Length $S$
        & KV Cache
        & Relative Memory \\
        \midrule
        256  & 128 MiB & $\sim 1.008\times$ \\
        512  & 256 MiB & $\sim 1.016\times$ \\
        1024 & 512 MiB & $\sim 1.034\times$ \\
        2048 & 1 GiB   & $\sim 1.067\times$ \\
        \bottomrule
    \end{tabular}
\end{table}

\section{Decoding Signal Definitions}
\label{app:signal_definitions}

This appendix provides the detailed definitions of the decoding signals introduced in Section~\ref{sec:tracequery_inference}. Let $i\in\mathcal{M}_t$ denote a still-masked candidate position and let $\mathcal{R}_t\subseteq\mathcal{U}_t$ denote the set of recently unmasked positions.

\paragraph{Confidence.}
Given the token probability distribution at position $i$, confidence is defined as
\begin{equation}
    c_i^{(t)}
    =
    \max_v
    p_{\theta}\left(v\mid x^{(t)}\right)_i.
\end{equation}

\paragraph{Probability margin.}
Let $p_{i,(1)}^{(t)}$ and $p_{i,(2)}^{(t)}$ denote the largest and second-largest token probabilities at position $i$. The probability margin is
\begin{equation}
    m_i^{(t)}
    =
    p_{i,(1)}^{(t)} - p_{i,(2)}^{(t)}.
\end{equation}

\paragraph{Mask density.}
For a local neighborhood $\mathcal{W}_i$ centered at candidate $i$, mask density is defined as
\begin{equation}
    \rho_i^{(t)}
    =
    \frac{1}{|\mathcal{W}i|}
    \sum{j\in\mathcal{W}_i}
    \mathbf{1}\left[j\in\mathcal{M}_t\right].
\end{equation}

\paragraph{Relative distance.}
Let $L$ be the generation block size. The distance from candidate $i$ to recently unmasked positions is defined as
\begin{equation}
    d_i^{(t)}
    =
    \frac{\min_{j\in\mathcal{R}_t}|i-j|}{L},
\end{equation}
and is normalized by the active generation length when required.

\paragraph{Last-layer attention.}
Let $A_{\ell,h}^{(t)}(j,i)$ denote the attention from position $j$ to position $i$ at layer $\ell$ and head $h$. The last-layer attention signal is
\begin{equation}
    a_{i,\mathrm{last}}^{(t)}
    =
    \operatorname{Agg}_{j\in\mathcal{R}t,\,h}
    A{L,h}^{(t)}(j,i),
\end{equation}
where $L$ denotes the final Transformer layer.

\paragraph{Full-layer attention.}
The full-layer attention signal aggregates the same dependency across Transformer layers:
\begin{equation}
    a_{i,\mathrm{full}}^{(t)}
    =
    \operatorname{Agg}_{\ell,\,j\in\mathcal{R}t,\,h}
    A{\ell,h}^{(t)}(j,i).
\end{equation}

\section{Inference and Training algorithm}
\input{algorithms/train}
\input{algorithms/inference}

%% file: algorithms/train.tex
\begin{algorithm}[h]
\caption{\sysname Router Training}
\label{alg:router_training}
\begin{algorithmic}[1]
\Require Training samples $\mathcal{X}$, diffusion model $p_\theta$,
         router $f_\phi$, query horizon $H$

\State $\mathcal{D} \gets \emptyset$

\ForAll{$x\in\mathcal{X}$}
    \State Run full-denoising decoding with one token unmasked per step
    \State Record teacher trajectory
        $\tau=(i_0,i_1,\ldots,i_{T-1})$
    \For{$t=0,\ldots,T-2$}
        \State Collect signals for all
            $i\in\mathcal{M}_{t+1}$
        \State Normalize last-layer attention with \textsc{Softmax}
        \State Store
            $\{\mathbf{z}_i^{(t)},g_i^{(t)}\}_{i\in\mathcal{M}_{t+1}}$
            in $\mathcal{D}$
    \EndFor
\EndFor

\Repeat
    \State Sample a decoding state $t$ from $\mathcal{D}$
    \ForAll{$i\in\mathcal{M}_{t+1}$}
        \State $s_i^{(t)} \gets f_\phi(\mathbf{z}_i^{(t)})$
    \EndFor

    \State $\mathcal{L} \gets 0$
    \For{$j=1,\ldots,H$}
        \State $\mathcal{C}_{t,j}
        \gets
        \{i\in\mathcal{M}_{t+1}:g_i^{(t)}\ge j\}$
        \State $i_{(j)} \gets$ teacher's $j$-th future unmask
        \State $\mathcal{L} \gets \mathcal{L}
        - \log
        \dfrac{
            \exp(s_{i_{(j)}}^{(t)})
        }{
            \sum_{i\in\mathcal{C}_{t,j}}\exp(s_i^{(t)})
        }$
    \EndFor

    \State Update $\phi$ using $\nabla_\phi \mathcal{L}$
\Until{convergence}

\State \Return trained router $f_\phi$
\end{algorithmic}
\end{algorithm}

%% file: algorithms/inference.tex
\begin{algorithm}[h]
\caption{\sysname Inference}
\label{alg:sparsequery_inference}
\begin{algorithmic}[1]
\Require Diffusion model $p_\theta$, router $f_\phi$, query horizon $H$,
         refresh schedule $\mathcal{R}$
\State Initialize masked set $\mathcal{M}_0$ and perform a full forward pass
\State Cache all KV states; set $\mathcal{Q}_0 \gets \mathcal{M}_0$

\For{$t=0,1,\ldots$ \textbf{while} $\mathcal{M}_t \neq \emptyset$}
    \If{$t \in \mathcal{R}$}
        \State $(P_t,A_t,\mathrm{KV}) \gets
               \Call{FullForward}{\mathcal{M}_t,\mathrm{KV}}$
        \State $\mathcal{E}_t \gets \mathcal{M}_t$
    \Else
        \State $(P_t,A_t,\mathrm{KV}) \gets
               \Call{QueryForward}{\mathcal{Q}_t,\mathrm{KV}}$
        \State $\mathcal{E}_t \gets \mathcal{Q}_t$
    \EndIf

    \State $i_t \gets
        \arg\max_{i\in\mathcal{E}_t}
        \Call{Confidence}{P_t(i)}$
    \State Commit position $i_t$ and update
        $\mathcal{M}_{t+1}\gets\mathcal{M}_t\setminus\{i_t\}$

    \If{$\mathcal{M}_{t+1}=\emptyset$}
        \State \textbf{break}
    \EndIf

    \ForAll{$i\in\mathcal{M}_{t+1}$}
        \State $\mathbf{z}_i^{(t)}
        \gets
        [a_{i,\mathrm{last}}^{(t)},\rho_i^{(t)},d_i^{(t)}]$
        \State $s_i^{(t)} \gets f_\phi(\mathbf{z}_i^{(t)})$
    \EndFor
    \State $\mathcal{Q}_{t+1}
        \gets
        \Call{TopH}{\{s_i^{(t)}\}_{i\in\mathcal{M}_{t+1}},H}$
\EndFor
\State \Return generated sequence
\end{algorithmic}
\end{algorithm}